\documentclass[10pt,twocolumn,letterpaper]{article}

\usepackage{cvpr}
\usepackage{times}
\usepackage{epsfig}
\usepackage{graphicx}
\usepackage{amsmath}
\usepackage{amssymb}

\usepackage{caption}
\usepackage{subcaption}
\usepackage{enumitem}
\usepackage{color}

\usepackage{hyperref}

\cvprfinalcopy 

\def\cvprPaperID{1822} 

\ifcvprfinal\fi

\DeclareMathOperator*{\argmax}{arg\,max}
\DeclareMathOperator*{\argmin}{arg\,min}

\begin{document}

\title{Recovering 6D Object Pose and Predicting Next-Best-View in the Crowd}

\author{Andreas Doumanoglou$^{1,2}$, Rigas Kouskouridas$^1$, Sotiris Malassiotis$^2$, Tae-Kyun Kim$^1$\\
\\
$^1$Imperial College London\\
$^2$Center for Research \& Technology Hellas (CERTH)\\
}

\maketitle

\newcommand{\LRF}{Latent Regression Forests}
\begin{abstract}

Object detection and 6D pose estimation in the crowd (scenes with multiple object instances, severe foreground occlusions and background distractors), has become an important problem in many rapidly evolving technological areas such as robotics and augmented reality. Single shot-based 6D pose estimators with manually designed features are still unable to tackle the above challenges, motivating the research towards unsupervised feature learning and next-best-view estimation.
In this work, we present a complete framework for both single shot-based 6D object pose estimation and next-best-view prediction based on Hough Forests, the state of the art object pose estimator that performs classification and regression jointly.  
Rather than using manually designed features we a) propose an unsupervised feature learnt from depth-invariant patches using a Sparse Autoencoder and b) offer an extensive evaluation of various state of the art features. Furthermore, taking advantage of the clustering performed in the leaf nodes of Hough Forests, we learn to estimate the reduction of uncertainty in other views, formulating the problem of selecting the next-best-view. To further improve pose estimation, we propose an improved joint registration and hypotheses verification module as a final refinement step to reject false detections. We provide two additional challenging datasets inspired from realistic scenarios to extensively evaluate the state of the art and our framework. One is related to domestic environments and the other depicts a bin-picking scenario mostly found in industrial settings. We show that our framework significantly outperforms state of the art both on public and on our datasets.

\end{abstract}

\vspace{-10px}
\section{Introduction}
\vspace{-4px}
Detection and pose estimation of everyday objects is a challenging problem arising in many practical applications, such as robotic manipulation~\cite{kouskouridas2014sparse}, tracking and augmented reality. Low-cost availability of depth data facilitates pose estimation significantly, but still one has to cope with many challenges such as viewpoint variability, clutter and occlusions. When objects have sufficient texture, techniques based on key-point matching \cite{martinez2010moped,Abbeel2012} demonstrate good results, yet when there is a lot of clutter in the scene they depict many false positive matches which degrades their performance. Also, holistic template-based techniques provide superior performance when dealing with texture-less objects 
\cite{hinterstoisser2011multimodal}, but suffer in cases of occlusions and changes in lighting conditions, while the performance also degrades when objects have not significant geometric detail. In order to cope with the above issues, a few approaches use patches \cite{tejani2014latent} or simpler pixel based features  \cite{brachmann2014learning} along with a Random Forest classifier. Although promising, these techniques rely on manually designed features which are difficult to make discriminative for the large range of everyday objects. 
\begin{figure}
\begin{center}
\includegraphics[width=0.9\linewidth]{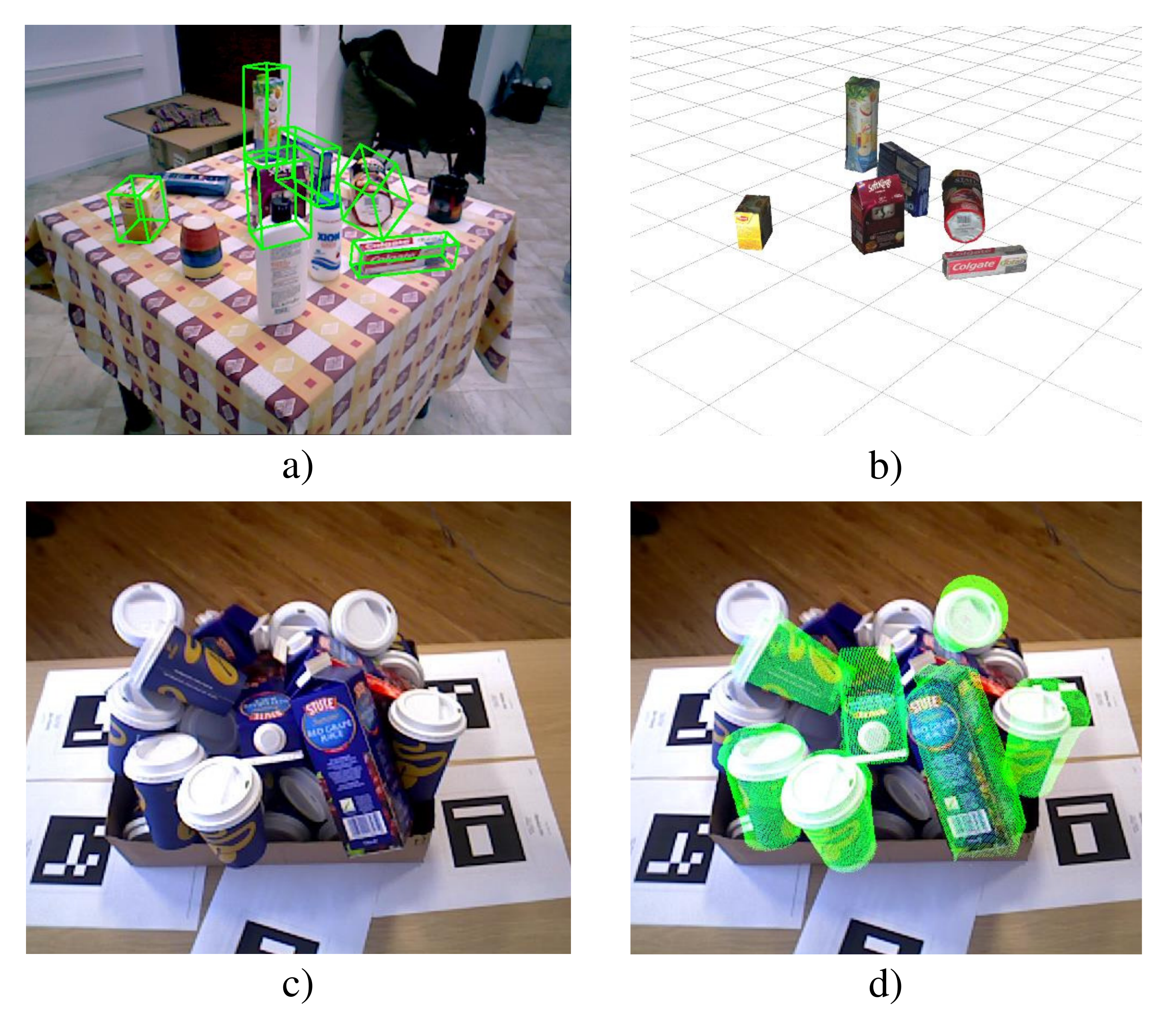}
\end{center}
\vspace{-12px}
   \caption{Sample photos from our dataset. a) Scene containing objects from a supermarket, b) our system's evaluation on a), c) Bin-picking scenario with multiple objects stacked on a bin, d) our system's evaluation on c).}
\label{fig:fig1}
\vspace{-15px}
\end{figure}
Last, even when the above difficulties are partly solved, multiple objects present in the scene, occlusions and distructors can make the detection very challenging from a single viewpoint, resulting in many ambiguous hypotheses. When the setup permits, moving the camera can be proved very beneficial for accuracy increase. The problem is how to select the next best viewpoint, which is crucial for fast scene understanding.

The above observations motivated us to introduce a complete framework for both single shot-based 6D object pose estimation and next-best-view prediction in a unified manner based on Hough Forests, a variant of Random Forest that performs classification and regression jointly \cite{tejani2014latent}. We adopted a patch-based approach but contrary to \cite{hinterstoisser2011multimodal,tejani2014latent,brachmann2014learning} we learn features in an unsupervised way using deep Sparse Autoencoders. The learnt features are fed to a Hough Forest \cite{gall2011hough} to determine object classes and poses using 6D Hough voting. To estimate the next-best-view, we exploit the capability of Hough Forests to calculate the hypotheses entropy, i.e. uncertainty, at leaf nodes. Using this property we can predict the next-best-viewpoint based on current view hypotheses through an object-pose-to-leaf mapping. We are also taking into account the various occlusions that may appear from the other views during the next-best-view estimation. Last, for further false positives reduction, we introduce an improved joint optimization step inspired by \cite{verification2012}. To the best of our knowledge, there is no other framework jointly tackling feature learning, classification, regression and clustering (for next-best-view) in a patch-based inference strategy.

In order to evaluate our framework, we do an extensive evaluation for single shot detection of various state of the art features and detection methods, showing that the proposed approach demonstrates a significant improvement compared to the state of the art techniques, on many challenging publicly available datasets. We also evaluate our next-best-view selection to various baselines and show its improved performance, especially in cases of occlusions. To demonstrate more explicitly the advantages of our framework, we provide an additional dataset consisting of two realistic scenarios shown in Fig. \ref{fig:fig1}. Our dataset also reveals the weaknesses of the state of the art techniques to generalize to realistic scenes.
In summary, our main contributions are:
\vspace{-8px}
\begin{itemize}[leftmargin=*]
\item A complete framework for 6 DoF object detection that comprises of a) an architecture based on Sparse Autoencoders for unsupervised feature learning, b) a 6D Hough voting scheme for pose estimation and c) a novel active vision technique based on Hough Forests for estimating the next-best-view.\vspace{-8px}
\item Extensive evaluation of features and detection methods on several public datasets. \vspace{-8px}
\item A new dataset of RGB-D images reflecting two usage scenarios, one representing domestic environments and the other a bin-picking scenario found in industrial settings. We provide 3D models of the objects and, to the best of our knowledge, the first fully annotated bin-picking dataset.
\end{itemize}

\vspace{-10px}
\section{Related Work}
\vspace{-4px}
Unsupervised feature learning has recently received the attention of the computer vision community. Hinton \etal \cite{Hinton2006} used a deep network consisting of Restricted Boltzmann Machines 
for dimensionality reduction and showed that deep networks can converge to a better solution by greedy layer-wise pre-training. Jarrett \etal \cite{LeCun2009} showed the merits of multi-layer feature extraction with pooling and local contrast normalization over single-layer architectures, while Le \etal \cite{icml12} used a 9-layer Sparse Autoencoder to learn a face detector only from unsupervised data. Feature learning has also been used for classification\cite{socher2012convolutional} using RNNs, and detection\cite{bo2014learning} using sparse coding, trained with holistic object images and patches, respectively. Coates \etal \cite{Coates2010}  investigated different single-layer unsupervised architectures such as k-means, Gaussian mixture modes, and Sparse Autoencoders achieving state of the art results when parameters were fine-tuned.
Here, we use the Sparse Autoencoders of \cite{Coates2010} but in a deeper network architecture, extracting features from raw RGB-D data.
In turn, in \cite{han2015matchnet} and \cite{zagoruyko2015learning} it was shown how CNNs could be trained for supervised feature learning, while in \cite{ouyang2015deepid} and \cite{riegler2013hough} CNNs were trained to perform classification and regression jointly for 2D object detection and head pose estimation, respectively.

Object detection and 6 DoF pose estimation is also frequently addressed in the literature. Most representative are techniques based on template matching, like LINEMOD~\cite{hinterstoisser2011multimodal}, its extension~\cite{Rios_Cabrera_2013_ICCV} and the Distance Transform approaches~\cite{liu2012fast}. Point-to-Point methods \cite{drost2010model,rusu2009fast} form another representative category where emphasis is given on building point pair features to construct object models based on point clouds. Tejani \etal \cite{tejani2014latent} combined Hough Forest with~\cite{hinterstoisser2011multimodal} using a template matching split function to provide 6 DoF pose estimation in cluttered environments. They provided evidence that, using patches instead of the holistic image of the object, can boost the performance of the pose estimator in cases of severe occlusions and clutter. Brachmann \etal \cite{brachmann2014learning} introduced a new representation in form of a joint 3D object coordinate and class labelling, which, however suffers in cases of occlusions. Additionally, Song \etal \cite{song2014sliding} proposed a computationally expensive approach to the 6 DoF pose estimation problem that slides exemplar SVMs in the 3D space, while in \cite{bonde2014robust} shape priors are learned by soft labelling Random Forest for 3D object classification and pose estimation. Lim \etal \cite{lim2014fpm} achieved fine pose estimation by representing geometric and appearance information as a collection of 3D shared parts and objectness, respectively. Wu \etal \cite{3dShap_Net} designed a model that learns the joint distribution of voxel data and category labels using a Convolutional Deep Belief Network, while the posterior distribution for classification is approximated by Gibbs sampling. The authors in \cite{wohlhart2015learning} tackle the 3D object pose estimation problem by learning discriminative feature descriptors via a CNN and then passing them to a scalable Nearest Neighbor method to efficiently handle a large number of objects under a large range of poses. However, compared to our work, this method is based on holistic images of the objects, which is prone to occlusions \cite{tejani2014latent} and only evaluated on a public dataset that contains no foreground occlusions.

Hypotheses verification is employed as a final refinement step to reject false detections. Aldoma \etal~\cite{verification2012} proposed a cost function-based optimization to increase true positive detections. Fioraio \etal~\cite{fioraio2013joint} showed how single-view hypotheses verification can be extended to multi-view ones in order to facilitate SLAM through a novel Bundle adjustment framework. Buch \etal~\cite{buch2014search} presented a two-stage voting procedure for estimating the likelihood of correspondences, within a set of initial hypotheses, between two 3D models corrupted by false positive matches. 

Regarding active vision, a recent work presented by Jia et al. \cite{Jia10} makes use of the Implicit Shape Model combined in a boosting algorithm to plan the next-best-view for 2D object recognition, while Atanasov \etal \cite{Atanasov14} proposed a non-myopic strategy using POMDPs for 3D object detection. Wu \etal \cite{3dShap_Net} used their generative model based on the convolutional network to plan for the next-best-view but is limited in the sense that the holistic image of the object is needed as input. Since previous works are largely dependent on the employed classifier, more related to our work is the recently proposed Active Random Forests \cite{andoum2014} framework, which, however (similar to \cite{3dShap_Net}) requires the holistic image of an object to make a decision, making it not appropriate for our patch-based method. 

\vspace{-6px}
\section{6 DoF Object Pose \& Next-Best-View Estimation Framework}
\vspace{-4px}
Our object detection and pose estimation 
framework consists of two main parts: a) single shot-based 6D object detection and b) next-best-view estimation. 
In the first part, we render the training objects and extract depth-invariant RGB-D patches. The latter are given as input to a Sparse Autoencoder which learns a feature vector in an unsupervised manner. Using this feature representation, we train a Hough Forest to recognize object patches in terms of class and 6D pose (translation and rotation). Given a test image, patches from the scene pass through the Autoencoder followed by the Hough forest, where the leaf nodes cast a vote in a 6D Hough space indicating the existence of an object. The modes of this space represent our best object hypotheses. The second part, next-best-view estimation, is based on the previously trained forest. Using the training sample distribution in the leaf nodes, we are able to determine the uncertainty, i.e. the entropy, of our current hypotheses, and further estimate the reduction in entropy when moving the camera to another viewpoint using a pose-to-leaf mapping. Fig. \ref{fig:pipeline} shows an overview of the framework. In the following subsections, we describe each part in detail.

\subsection{Single Shot-based 6D Object Detection}
\vspace{-2px}

\noindent \textbf{State of the art Hough Forests Features} In the literature some of the most recent 6D object detection methods use Hough Forests as their underlying classifier. In \cite{brachmann2014learning} simple two pixel comparison tests were used to split the data in the tree nodes, while the location of the pixels could be anywhere inside the whole object area. In our experiments, we also added the case where the pixel tests are restricted inside the area of an image patch. A more sophisticated feature for splitting the samples was proposed by Tejani et al. \cite{tejani2014latent} who used a variant of the template based LineMOD feature \cite{hinterstoisser2011multimodal}. In comparison with the above custom-designed features, we use Sparse Autoencoders to learn an unsupervised feature representation of varying length and layers. Furthermore, we learn features over depth-invariant RGB-D patches extracted from the objects, as described below.\vspace{2px}

\begin{figure*}
\begin{center}
\includegraphics[width=0.9\linewidth]{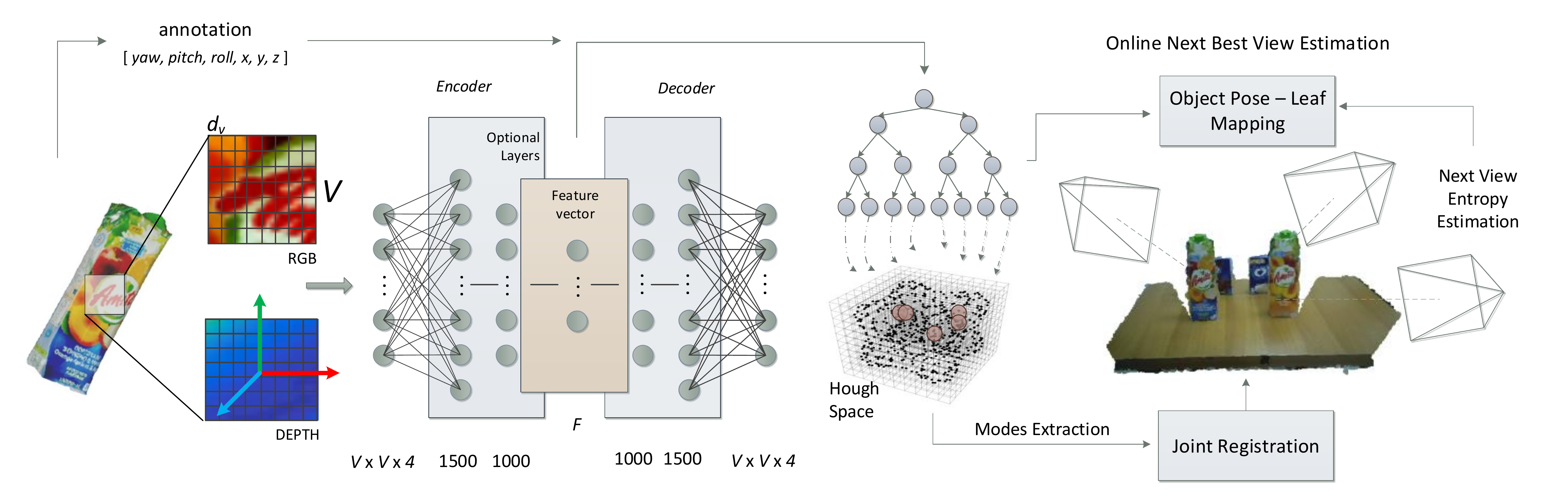}
\end{center}
\vspace{-12px}
   \caption{Framework Overview. After patch extraction, RGBD channels are given as input to the Sparse Autoencoder. The annotation along with the produced features of the middle layer are given to a Hough Forest, and the final hypotheses are generated as the modes of the Hough voting space. After refining the hypotheses using joint registration, we estimate the next-best-view using a pose-to-lead mapping learnt from the trained Hough Forest.}
\label{fig:pipeline}
\vspace{-15px}
\end{figure*}

\noindent\textbf{Patch Extraction}\label{patch_extraction} Our approach relies on 3D models of the objects of interest.
We render synthetic training images by placing a virtual camera on discrete points on a sphere surrounding the object. In traditional patch-based techniques \cite{gall2011hough}, the patch size is expressed directly in image pixels. In contrast, we want to extract depth invariant, 2.5D patches that cover the same area of the object regardless of the object distance from the camera, similar to \cite{Sun2010}. First, a sequence of patch centers $c_i, i=1..N$ is defined on a regular grid on the image plane. Using the depth value of the underlying pixels these are back-projected to the 3D world coordinate frame, i.e. $\overline{c}_i = (x,y,z)$. For each such 3D point $\overline{c}_i$ we define a planar patch perpendicular to the camera, centered at $\overline{c}_i$ and with dimensions $d_p \times d_p$, measured in meters, which is subdivided into $V \times V$ cells. Then, we back-project the center of each cell to the corresponding point on the image plane, to compute its RGB and depth values via linear interpolation\footnote{The cell values calculation can be done efficiently and in parallel using texture mapping in gpu.}. Depth values are expressed with respect to the frame centered at the center of the patch (Fig. \ref{fig:pipeline}). Also, we truncate depth values to a certain range to avoid points not belonging to the object. Depth-invariance is achieved by expressing the patch size in metric units in 3D space. From each training image we extract a collection of patches $\mathbf{P}$ and normalize their values to the range $[0, 1]$. The elements corresponding to the four channels of the patch are then concatenated into a vector of size $V \times V \times 4$ (RGBD channels) and are given as input to the Sparse Autoencoder for feature extraction. \vspace{2px}

\noindent\textbf{Unsupervised Feature Learning} We learn unsupervised features using a network consisting of stacked, fully connected Sparse Autoencoders, in a symmetric encoder-decoder scheme. An autoencoder is a fully connected, symmetric neural network, that learns to reconstruct its input. If the number of hidden units are limited or a small number of active units is allowed (sparsity), it can learn meaningful representations of the data. In the simplest case of one hidden layer with $F$ units, one input ($x$) and one output ($y$) layer of size $N$, the Autoencoder finds a mapping $f: \mathbb{R}^N \rightarrow \mathbb{R}^F$ of the input vectors $x \in \mathbb{R}^N$ as: 
\begin{equation}
f = sigm(Wx+b)
\end{equation}
The weights $W \in \mathbb{R}^{F\times N}$ and the biases $b \in \mathbb{R}^F$ are optimized by back-propagating the reconstruction error ${|| y - x ||_2}$. The average activation of each hidden unit is enforced to be equal to $\rho$, a sparsity parameter with a value close to zero. The mapping $f$ represents the features given as input to the classifier in the next stage. We can extract ``deeper'' features by stacking several layers together, to form an encoder-decoder symmetric network as shown in Fig. \ref{fig:pipeline}. In this case, the features are extracted from the last layer of the encoder (i.e. middle layer). In experiments, we use one to three layers in the encoder part, and analyse the effect of several parameters of the architecture on the pose estimation performance, such as the number of layers, the number of features and layer-wise pre-training \cite{Hinton2006}.\vspace{2px}

\noindent\textbf{Pose Estimation} During training, we extract patches from training images of objects and use the trained network to extract features from object patches, that form a feature vector $\mathbf{f}=\{f_1, f_2, ..., f_F\}$. These vectors are annotated using a vector $\mathbf{d}$ that contains the object class, the pose of the object in the training image and the coordinates of the patch center expressed in the object's frame, i.e. $\mathbf{d}=\{class, yaw, pitch, roll, x, y, z\}$. The feature vectors along with their annotation are given as input to the Hough Forest. We propose three different objective functions: entropy minimization of the class distribution of the samples, entropy minimization of the $\{yaw,pitch,roll\}$ variables, and entropy minimization of the $\{x,y,z\}$ variables. Reducing the entropy towards the leaves, has the effect of clustering the training samples that belong to the same class and having similar position and pose on the object. More details on the computation of these entropies can be found in \cite{andoum2014,Criminisi2011TR}. The objective function used is randomly selected in each internal node and samples are split using axis aligned random tests. The leaf nodes store a histogram of the observed classes of the samples that arrived, and a list of the annotation vectors $\mathbf{d}$. 
During testing we extract patches from the test image with a stride $s$ and pass them through the forest, to reach the corresponding leaf. We create a separate Hough voting space (6D space) for each object class, where we accumulate the votes of the leaf nodes. Each vector $\mathbf{d}$ stored in the leafs, casts a vote for the object pose and its center to the corresponding Hough space. The votes are weighted according to the probability of the associated class stored in the leaf. Object hypotheses are subsequently obtained by estimating the modes of each Hough space. Each mode can be found using non-maxima suppression and is assigned a score equal to the voting weight of the mode.


\subsection{Next-Best-View Prediction}
\vspace{-2px}

\begin{figure*}
\begin{center}
\includegraphics[width=1\linewidth]{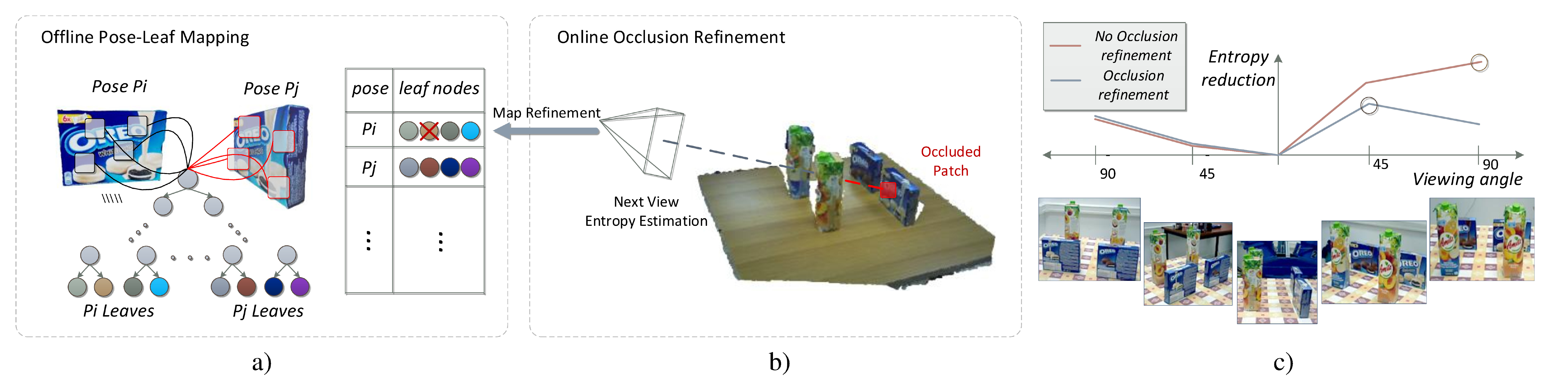}
\end{center}
\vspace{-12px}
   \caption{a) Offline construction of the pose-to-leaf mapping, b) Online occlusion refinement of the mapping, c) example of the effect of occlusion refinement in entropy estimation. }
\label{fig:occ_ref}
\vspace{-15px}
\end{figure*}

When detecting static objects, next-best-view selection is often achieved by finding the viewpoint that minimizes the expected entropy, i.e the uncertainty of the detection in the new viewpoint. There have been various methods proposed for computing the entropy reduction \cite{Atanasov14,andoum2014,3dShap_Net}. Hough Forests can facilitate the process since they store adequate information in the leaf nodes that can be used for predicting such reduction. The entropy of a hypothesis in the current view can be computed as the entropy of the samples stored in the leaf nodes that voted for this hypothesis. That is:
\vspace{-8px}
\begin{equation}
H(h) = \sum_{l_h}H(\mathbf{S}_{l_h})
\vspace{-6px}
\end{equation}
where $l_h$ is a leaf voted for hypothesis $h$, and $\mathbf{S}_{l_h}$ the set of samples in these leaves. 
If the camera moves to viewpoint $v$, the reduction in entropy we gain is:
\vspace{-8px}
\begin{equation}
\label{eq:reduction}
r(v) = H(h) - H(h_v) = \sum_{l_h}H(\mathbf{S}_{l_h}) - \sum_{l_{h_v}}H(\mathbf{S}_{l_{h_v}})
\vspace{-6px}
\end{equation}
where $h_v$ is the hypotheses $h$ as would be seen from viewpoint $v$.
In order to measure the reduction in entropy, we need to calculate the second term of the right side of equation \eqref{eq:reduction}, which requires to find the leaf nodes that should be reached from the viewpoint $v$. Since we want to compute the reduction before actually moving the camera, we can simulate $h_v$ by rendering the object placing a virtual camera at $v$, give the image as input to the forest and collect the resulting leaves. However this can be done more efficiently avoiding the rendering phase (contrary to \cite{3dShap_Net}): we save offline a mapping from object poses (discrete camera views) to leaf nodes using the training data as shown in Fig. \ref{fig:occ_ref}a. Given a 6 DoF hypothesis and this mapping, we can predict which leaf nodes of the forest are going to be reached if we move the camera to viewpoint $v$. Because of the discretization of poses in the map index, we choose the view in the mapping that is closer to the camera viewpoint we want to examine. Thus, the next-best-view $v_{best}$ is calculated as:
\vspace{-4px}
\begin{equation}
v_{best} = \argmax_v r(v) = \argmin_v H(h_v)
\vspace{-4px}
\end{equation}
In case of two or more uncertain hypotheses, the reduction in entropy is averaged in the new viewpoint. Also, to account for the cost of the movement, the reduction can be normalized by the respective cost.

In the general case of multiple objects present in the scene with cluttered background, we can further refine the entropy prediction to account for occlusions. In our previous formulations, we made the assumption that, from a view $v$ the object is clearly visible. However, due to other objects present in the scene, some part or the whole object we are interested in, may be occluded (Fig. \ref{fig:occ_ref}b). In this case our estimated entropy reduction is not correct. What we need to do is to exclude from the entropy calculation the samples in the leaves that are going to be occluded. More formally:
\vspace{-4px}
\begin{equation}
H(h_v) = \sum_{l_{h_v}}H(\mathbf{S}_{l_{h_v}} \setminus \mathbf{S}_{l_{h_v}}^{occ})
\vspace{-4px}
\end{equation}
where $\mathbf{S}_{l_{h_v}}^{occ}$ are the samples that would be occluded in viewpoint $v$. In order to determine this set, first, we incrementally update the 3D point cloud of the scene when the camera moves. Then, we project the $\{x, y, z\}$ coordinates of an annotated sample of a leaf onto the acquired scene as shown from view $v$, and estimate if it is going to be occluded or not. Figure \ref{fig:occ_ref}c shows an example of our dataset, where there are two similar objects \textit{oreo} we want to disambiguate. From the view -90 degrees to 0 it is difficult to understand the difference, while from 45 degrees onwards the difference becomes clearer. However, from the view of 90 degrees the objects of interest become occluded. Calculating the entropy as described above, we get the true reduction in entropy which is lower than in the 45 degree case.

\begin{figure*}[t]
\begin{center}
\includegraphics[width=0.9\linewidth]{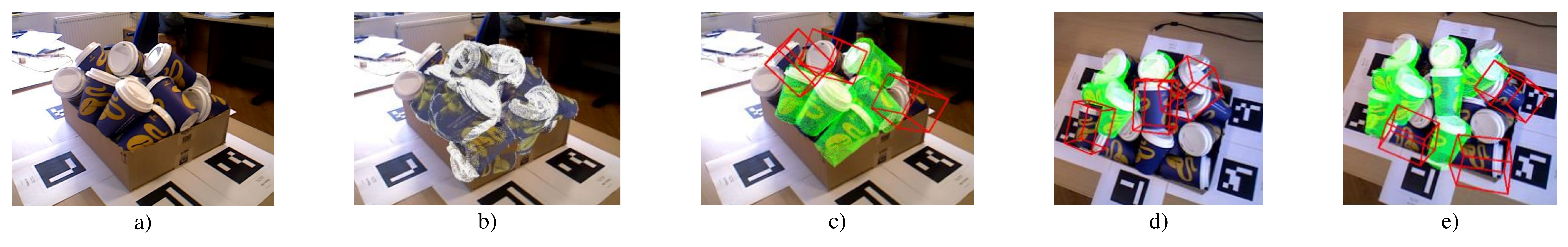}
\vspace{-12px}
\end{center}
   \caption{Example of hypotheses verification and active camera movement. a) Input test image, b) complete set of hypotheses overlaid on the image, c) hypotheses verification refinement, d) active camera movement, e) re-estimating hypotheses.}
\label{fig:verif_active}
\vspace{-15px}
\end{figure*}

Another example of the complete pipeline is shown in Fig. \ref{fig:verif_active}. Given an image (Fig. \ref{fig:verif_active}a) we extract the hypotheses from the Hough voting space (Fig. \ref{fig:verif_active}b). Using the optimization described in next section \ref{optimization} we refine this by selecting the best subset (Fig. \ref{fig:verif_active}c). The best solution does not include the objects shown in red box. However, a solution containing these hypotheses, but not well aligned with the scene due to occlusion, has a similar low cost with the best one. Being able to move the camera, we find the next-best-view as described above according to the uncertain hypotheses and change the viewpoint of the camera (Fig. \ref{fig:verif_active}d). We can re-estimate a new set of hypotheses (Fig. \ref{fig:verif_active}e) with some hypotheses still being uncertain (but keeping good ones above a threshold) and the same process is repeated. 

\subsection{Hypotheses verification and joint registration}
\label{optimization}
\vspace{-2px}
State of the art methods \cite{hinterstoisser2011multimodal,brachmann2014learning} assume that only one object instance exists in the scene. In case of multiple instances however, the produced set of hypotheses may be conflicting. To address this issue, we improved the global optimization approach of \cite{verification2012} in order to automatically select the subset of all possible hypotheses that best explains the scene. For each hypothesis we render the 3D model of the object in the scene, we exclude the parts that are occluded and define $p$ as a point of the object model and $q$ its nearest neighbor in the scene. If $||p-q||>p_e$ where $p_e$ a small constant, $p$ is considered an outlier. Given a set of hypotheses $\mathbf{H}$ and a vector $\mathbf{X} = \{x_1, .., x_i, .., x_N\}$ of boolean variables, which indicate that hypotheses $h_i$ is valid or not, we introduce the objective function $C(\mathbf{X})$ that should be minimized in terms of $\mathbf{X}$:
\vspace{-8px}
\begin{equation}
C(\mathbf{X}) = (a_1C_1 + a_2C_2) - (a_3C_3 + a_4C_4)
\vspace{-4px}
\end{equation}
where \\
$C_1=(C_{11}+C_{12}+C_{13})/3$\\
$C_{11}$: normalized distance $||p-q||/p_e$\\
$C_{12}$: dot product of the normals of the points\\
$C_{13}: max(\frac{|R_p - R_q|}{255}, \frac{|G_p - G_q|}{255}, \frac{|B_p - B_q|}{255})$ color similarity\\
$C_2 = p_{in} / p_{tot}$, $p_{in}$ points for which $||p - q|| \leq p_e$\\
$C_3$: fraction of conflicting inliers over the total inliers\\
$C_4$: fraction of penalized points over total points in a region\\
Each $C_i$ is calculated only for the hypotheses that $x_i = 1$. Contrary to \cite{verification2012}, we normalize every term in the range $[0, 1]$ and noted that each one has a different relative importance and common range of values. Therefore, unlike \cite{verification2012}, we put a different regularizer $a_i$ in each term, which is found using cross-validation. Furthermore, we further reduce the solution space of the optimization by splitting the set of hypotheses $\mathbf{H}$ into non-intersecting subsets $H_i$. Each subset can be optimized independently, decomposing the problem and reducing the time and complexity of the solution.

\section{Experiments}
\vspace{-2px}
The experiments regarding the patch size and feature evaluation were performed on a validation set of our own dataset. Object detection accuracy is measured using the F1-score and is averaged over the whole set of objects. When comparing with the state of the art methods, we use the public datasets and the evaluation metrics provided by the corresponding authors. When evaluating on our own dataset, we exclude the aforementioned evaluation set. \vspace{2px}

\begin{figure*}[t]
        \centering        
        \begin{subfigure}[t]{0.16\linewidth}
                \includegraphics[height=3.0cm]{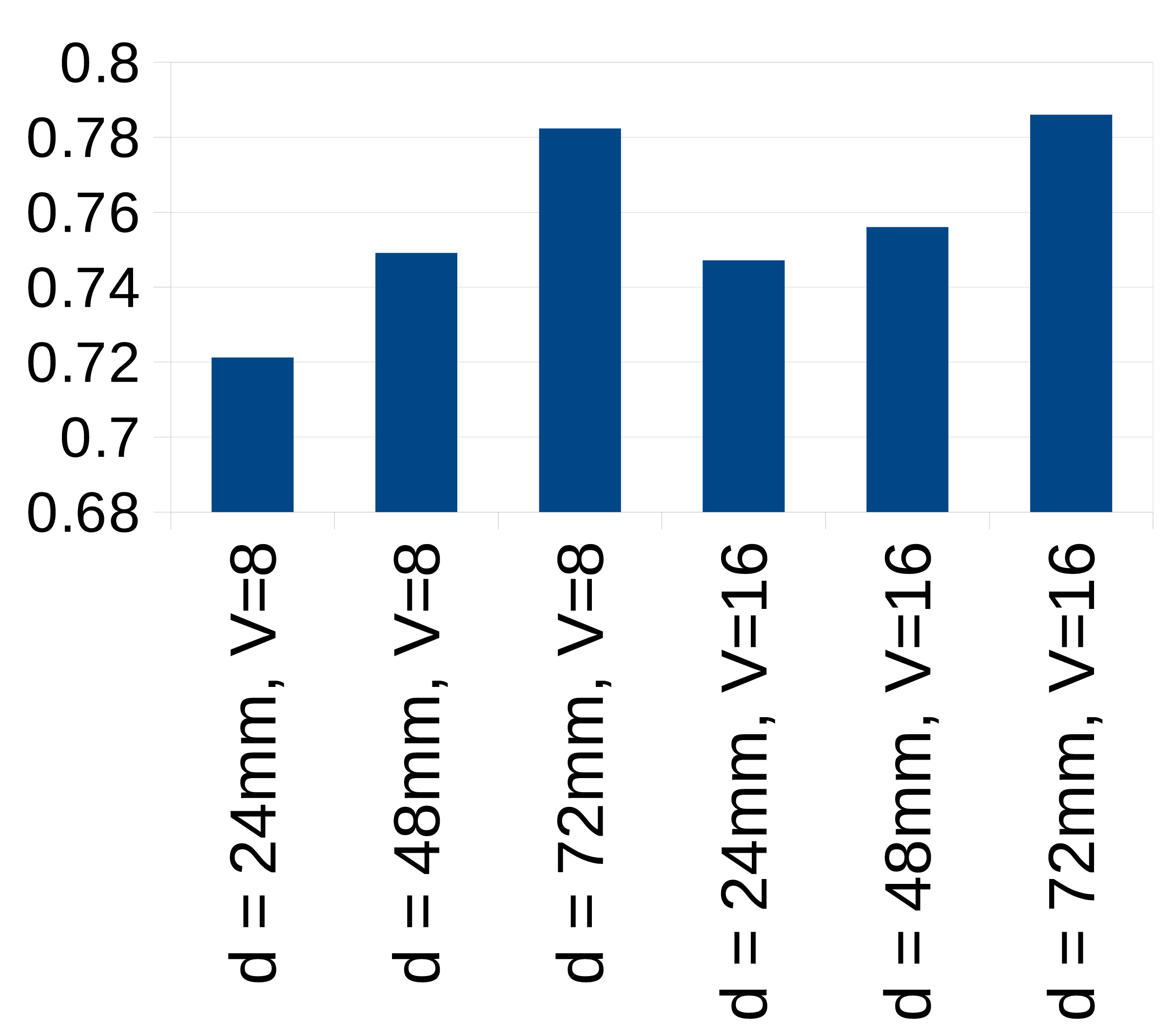}
                \caption{Patch-grid size}
                \label{fig:patch_size}
        \end{subfigure}\hfill%
        \begin{subfigure}[t]{0.13\linewidth}
                \includegraphics[height=3.0cm]{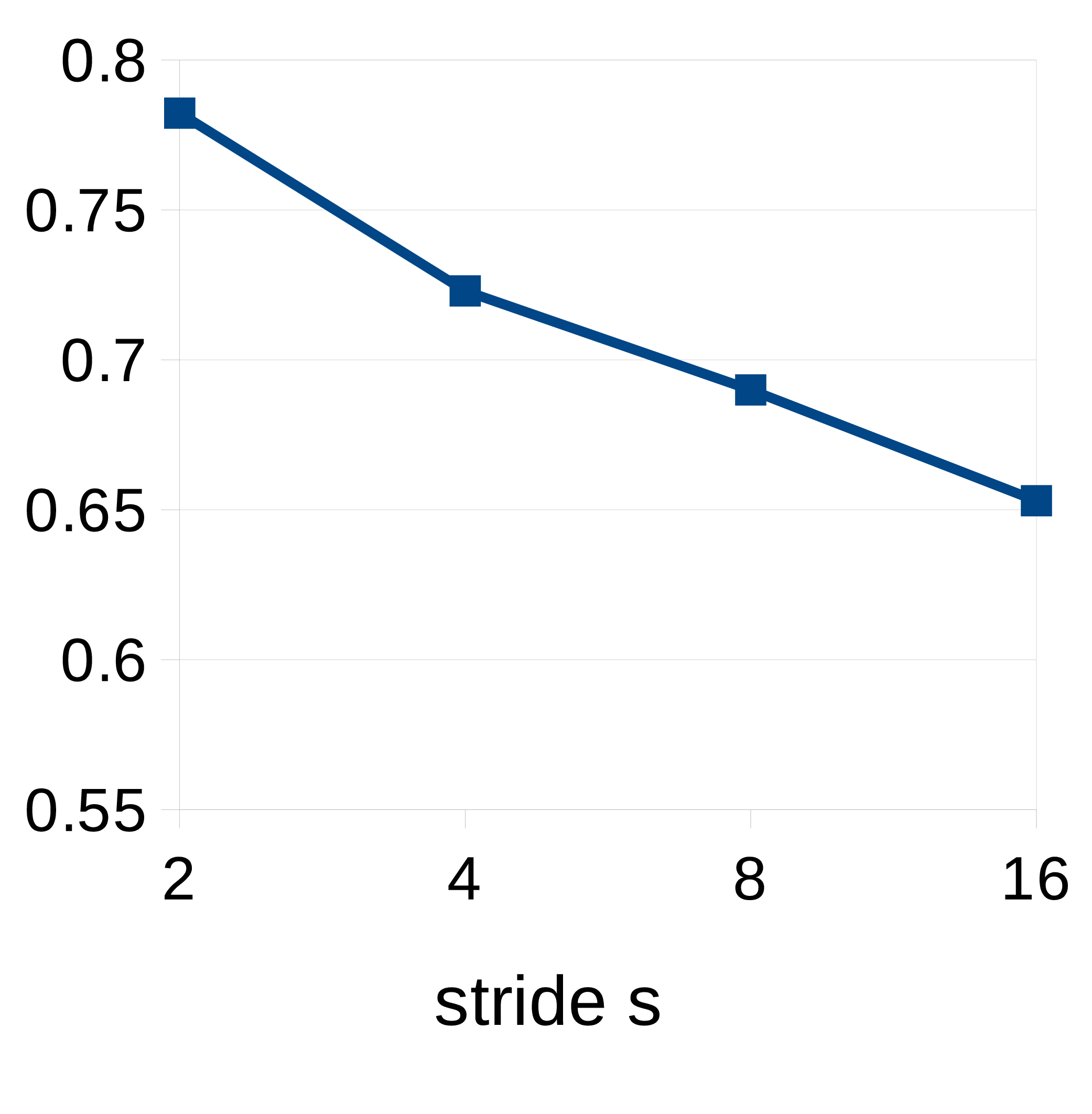}
                \caption{stride}
                \label{fig:stride}
        \end{subfigure}\hfill%
        \begin{subfigure}[t]{0.36\linewidth}
                \includegraphics[height=2.8cm]{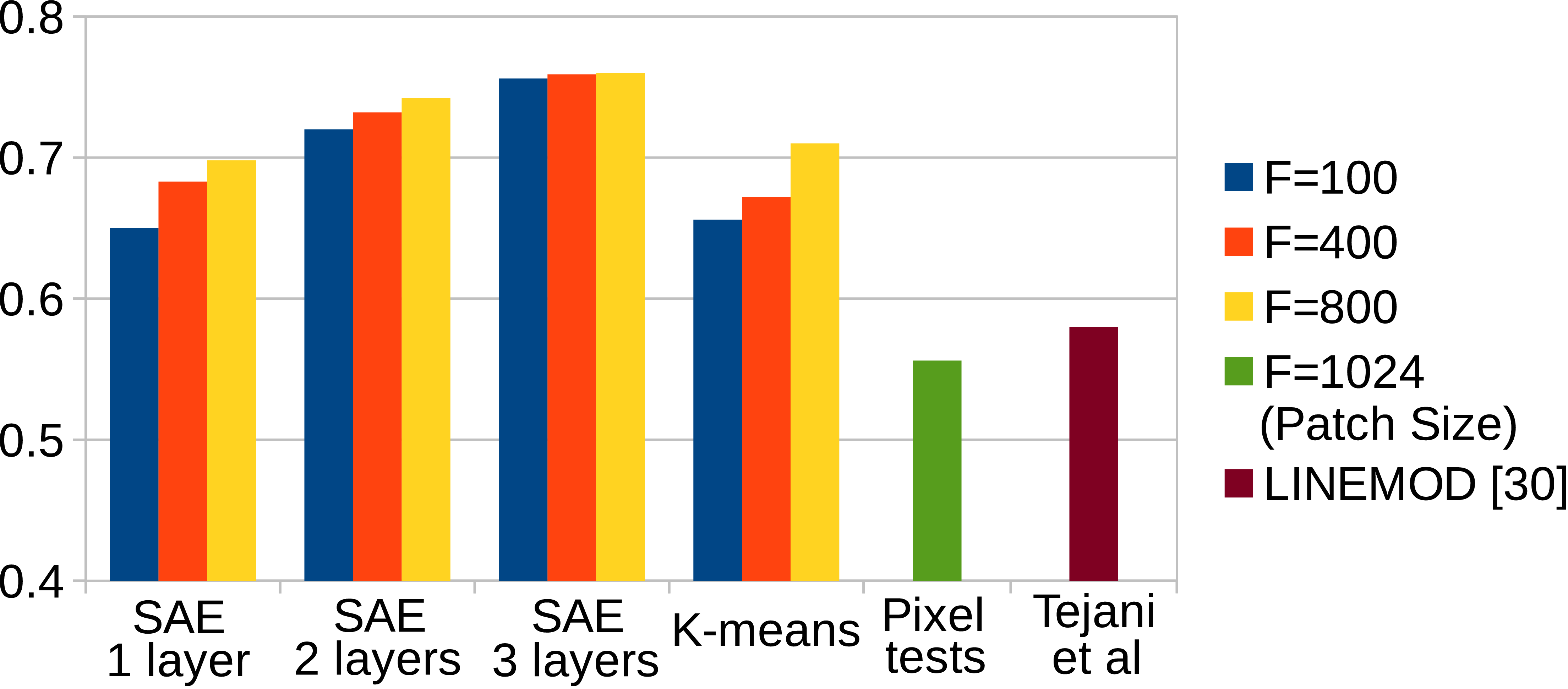}
                \caption{feature evaluation}
                \label{fig:layers-features}
        \end{subfigure}\hfill%
        \begin{subfigure}[t]{0.15\linewidth}                
                \raisebox{15px}{\includegraphics[height=2cm]{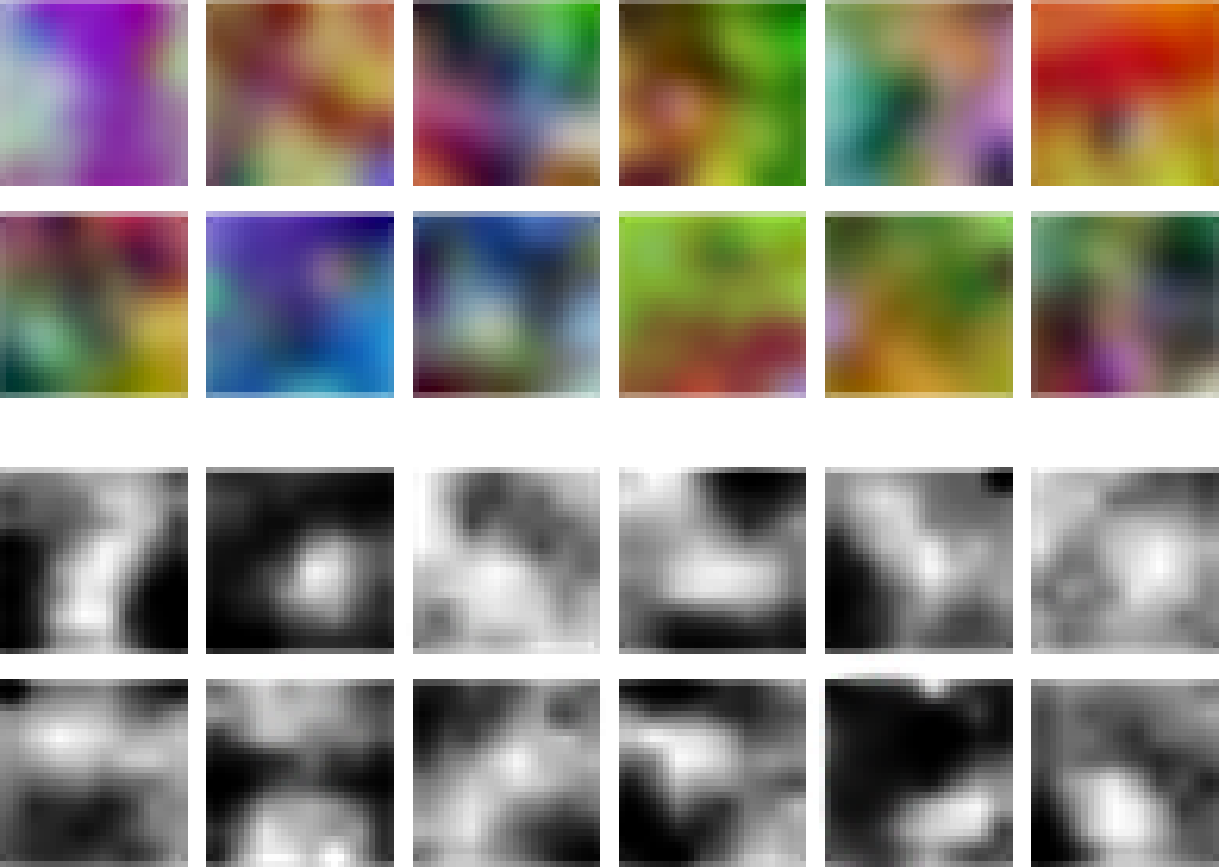}}
                \caption{1st layer filters}
                \label{fig:filters}
        \end{subfigure}
        \caption{Patch extraction parameters}\label{fig:patch-params}
        \vspace{-15px}
\end{figure*}

\noindent \textbf{Patch Size Evaluation} A patch in our framework is defined over 2 parameters: $d_p$ is the actual size measured in meters, and $V \times V$ is the number of cells a patch contains, which can be considered as the patch resolution. We used six different configurations shown in Fig. \ref{fig:patch_size}. 
The maximum patch size used was limited to the 2/3 of the smallest object dimensions. The network architecture used for patch-size experiments is 2 layers (the encoder part) of 1000 and 400 hidden units respectively. Fig. \ref{fig:patch_size} shows that an increase in the patch size significantly increases the accuracy, while on the other hand, an increase of the resolution offers a slight improvement, and that comes at the expense of additional computational cost. Another important factor is the stride $s$ during patch extraction. Fig. \ref{fig:stride} shows that the smaller the stride the more accurate the detection becomes. \vspace{2px}

\noindent \textbf{Feature Evaluation using Hough Forests} In order to evaluate our unsupervised feature we created 9 different network configurations to test the effect of both the number of features and the number of layers on the accuracy. We used 1-3 layers as the encoder of the network with the last layer of the encoder forming the feature vector used in the Hough Forest. We varied the length of this feature vector to be 100, 400 and 800. When we use 2 layers, the first has 1000 hidden units, while when we use 3 layers, the first two have 1500 and 1000 hidden units respectively. The patch size used for these experiments is $d_p = 48mm$ with $V=16$, creating an input vector of 1024 dimensions. Using the same Hough Forest configuration, we evaluate three state of the art features: a) the feature of \cite{tejani2014latent}, a variant of LineMOD\cite{hinterstoisser2011multimodal} designed for Hough Forests, along with its split function, b) the widely used pixel-tests \cite{brachmann2014learning} and c) K-means clustering, the unsupervised single-layer method that performed best in \cite{Coates2010}\footnote{We used the K-means (triangle) as described in \cite{Coates2010}} with 100, 400 and 800 clusters. Pixel-tests have been conducted inside the area of a patch for comparison purposes, however in the next subsection we compare the complete framework of \cite{brachmann2014learning} with ours. Results are shown in Fig. \ref{fig:layers-features}. The 3-layer Sparse Autoencoder shown the best performance. Regarding the Autoencoder, we notice that the accuracy increases if more features are used, but when the network becomes deeper, the difference diminishes. However, it can be seen that deeper features significantly outperform shallower ones. K-means performed slightly better than single-layer SAE, while pixel-tests had worse performance. The feature of \cite{tejani2014latent} had on average worse performance than Autoencoders and K-means, which is due to low performance on specific objects of the datasets. We further provide a visualization of the filters of the first layer learned by a network with a 3-layer encoder (Fig. \ref{fig:filters}). The first two rows are filters in the RGB channel, where it can be seen a bias towards the objects used for the evaluation. Filters in the depth channel resemble simple 3D edge and corner detectors. Last, we have tried to pre-train each layer as in \cite{Hinton2006}, without significantly influencing the results. \vspace{-9px}

\noindent \textbf{State of the Art Evaluation} In the experiments described in this subsection, we used an encoder of 3 layers with 1500, 1000 and 800 hidden units, respectively. The patch used has $V=8$ and $d_p=48mm$, which was found suitable for a variety of object dimensions. The forests contain four trees limiting only the number of samples per leaf to 30. For a fair comparison, we do not make use of joint registration or active vision except when specifically mentioned.

We tested our solution on the dataset of \cite{brachmann2014learning}, which contains 20 objects and a set of images regarded as background. The test scenes contain only one object per image, there is no occlusion or clutter, and are captured with different illumination from the training set, so one can check the generalization of a 6 DoF algorithm to different lighting conditions. To evaluate our framework we extracted the first $K=5$ hypotheses from the Hough voting space and chose the one with the best local fitting score. The results are shown in Table \ref{mic_tab} where for simplicity we show only 6 objects and the average over the complete dataset. Authors provided comparison with \cite{hinterstoisser2011multimodal} only with one object, because it was difficult to get results using their method.
This dataset was generally difficult to evaluate, mainly because some pose annotations were not very accurate, resulting in having some better estimations from the ground truth exceeding the metric threshold of acceptance. More details and results are included in the supplementary material.
Our method showed that it can generalize well on different lighting conditions, even without the need of modifying the training set with Gaussian noise as suggested by the authors. \vspace{-4pt}

\begin{table}[h]
\vspace{0px}
\fontsize{8}{9}\selectfont 
\centering
\caption{Results on the dataset of \cite{brachmann2014learning} (More on supplementary)}
\label{mic_tab}
\begin{tabular}{cccc}
\multicolumn{1}{l|}{Object}                       & \cite{hinterstoisser2011multimodal} (\%)    & \multicolumn{1}{l}{ \cite{brachmann2014learning} (\%)} & \multicolumn{1}{l}{\textbf{Our} (\%)} \\ \hline
\multicolumn{1}{l|}{Hole Puncher}     & -                         & \textbf{98.1}                             & 94.3                                  \\
\multicolumn{1}{l|}{Duck}             & -                         & 81.6                                      & \textbf{87.7}                         \\
\multicolumn{1}{l|}{Owl}              & -                         & 60.5                                      & \textbf{90.27}                        \\
\multicolumn{1}{l|}{Sculpture 1}      & -                         & 82.7                                      & \textbf{89.5}                         \\
\multicolumn{1}{l|}{Toy (Battle Cat)} & \multicolumn{1}{c}{70.2} & 91.8                                      & \textbf{92.4}  \\
\multicolumn{1}{l|}{...}      &          -                & ...                                      &                         \\
\hline
\multicolumn{1}{l|}{Avg.}             &   -                       & 88.2                                      & \textbf{89.1}                         \\
\end{tabular}
\vspace{-4px}
\end{table}

\begin{figure*}
        \centering        
        \begin{subfigure}[t]{0.14\textwidth}
                \includegraphics[height=2.25cm]{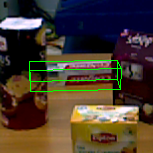}
                \caption{Colgate}
                \label{fig:colgate}
        \end{subfigure}
        \begin{subfigure}[t]{0.14\textwidth}
                \includegraphics[height=2.25cm]{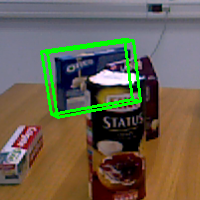}
                \caption{Oreo}
                \label{fig:oreo}
        \end{subfigure}
        \begin{subfigure}[t]{0.14\textwidth}
                \includegraphics[height=2.25cm]{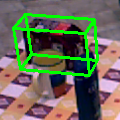}
                \caption{Softkings}
                \label{fig:softkings}
        \end{subfigure}
        \begin{subfigure}[t]{0.14\textwidth}
                \includegraphics[height=2.25cm]{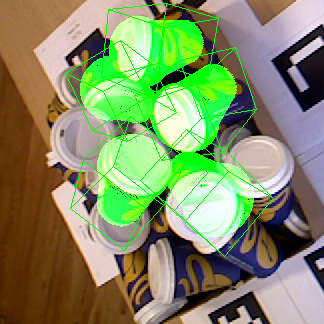}
                \caption{Coffecup}
                \label{fig:coffecup_bin}
        \end{subfigure}
        \begin{subfigure}[t]{0.14\textwidth}
                \includegraphics[height=2.25cm]{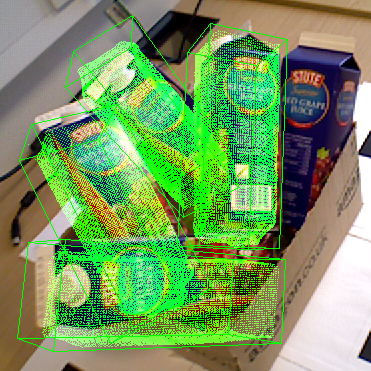}
                \caption{Juice}
                \label{fig:juice_bin}
        \end{subfigure}
        \begin{subfigure}[t]{0.14\textwidth}
                \includegraphics[height=2.25cm]{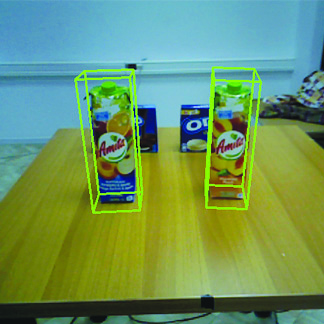}
                \caption{Camera}
                \label{fig:active1}
        \end{subfigure}
        \begin{subfigure}[t]{0.14\textwidth}
                \includegraphics[height=2.25cm]{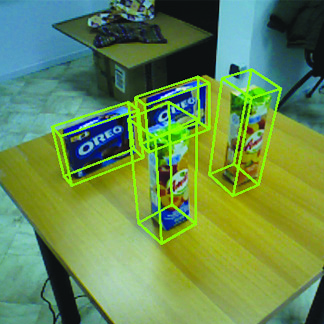}
                \caption{Joystick}
                \label{fig:active2}
        \end{subfigure}\hfill
        \caption{Qualitative results of our framework. Image \ref{fig:active2} is the next best view of image \ref{fig:active1}.}\label{fig:qualitative_results}
        \vspace{-8px}
\end{figure*}

We have also tested our method on the dataset presented in \cite{tejani2014latent}, which contains multiple objects of one category per test image, with much clutter and some cases of occlusion. Authors adopted one-class training, thus, avoiding background class images during training. For comparison, we followed the same strategy. 
Since there are multiple objects in the scene, we extract the top $K=10$ modes of the $\{x, y, z\}$ Hough space, and for each mode, we extract the $H=5$ modes of the $\{yaw, pitch, roll\}$ Hough space and put a threshold on the local fitting of the final hypotheses to produce the PR curves. Table \ref{aly_tab} shows the results in the form of F1-score (metric authors used) for each of the 6 objects. The results of methods \cite{hinterstoisser2011multimodal,drost2010model} are taken from \cite{tejani2014latent}.

\begin{table}[h]
\vspace{-4px}
\fontsize{8}{9}\selectfont 
\centering
\caption{Results on the dataset of \cite{tejani2014latent} }
\label{aly_tab}
\begin{tabular}{l|llll}
\textbf{Object} & \textbf{\cite{hinterstoisser2011multimodal}} & \textbf{\cite{drost2010model}} & \textbf{\cite{tejani2014latent}} & \textbf{Our} \\ \hline
                & \multicolumn{4}{c}{\textbf{F1 score}}               \\
Coffee Cup      & 0.819      & 0.867      & 0.877      & \textbf{0.932}              \\
Shampoo         & 0.625      & 0.651      & \textbf{0.759}      & 0.735             \\
Joystick        & 0.454      & 0.277      & 0.534      & \textbf{0.924}             \\
Camera          & 0.422      & 0.407      & 0.372      & \textbf{0.903}             \\
Juice Carton    & 0.494      & 0.604      & \textbf{0.870}      &  0.819            \\
Milk            & 0.176      & 0.259      & 0.385      &  \textbf{0.51}            \\ \hline
Average         & 0.498      & 0.511      & 0.633      &  \textbf{0.803}           
\end{tabular}
\vspace{-10px}
\end{table}

In this dataset we see that our method significantly outperforms the state of arts, especially regarding the \textit{Camera} which is small and looks similar with the background objects, and the Joystick, which has a thin and a thick part. Our features showed better performance on \textit{Milk} that contains other distracting objects on it.
It is evident that our learnt features are able to handle a variety of object appearances with stable performance and at the same time being robust to destructors and occluders. Note that without explicitly training a background class, all the patches in the image are classified as belonging to one of our objects. While \cite{tejani2014latent} designed a specific technique to tackle this issue, our features seem informative enough to produce good modes in the Hough spaces.

We have also tested \cite{tejani2014latent} and \cite{brachmann2014learning} on our own dataset. We also tried \cite{hinterstoisser2011multimodal}, but although we could produce the reported results on their dataset, we were not able to get meaningful results on our dataset and so we do not report them. This is mainly because this method is not intended to be used in textured objects with simple geometry. We provide results both with and without using joint object optimization. Our dataset contains 3D models of six training objects, while the test images may contain other objects as well. More on our dataset and evaluation can be found in the supplementary material.
Table \ref{tab:our_db} shows the results on our database. The work of \cite{brachmann2014learning} is designed to work only with one object per image and it is not evaluated 
on the bin-picking dataset. Our method outperforms all others even without joint optimization, but we can clearly see the advantages of such optimization on the final performance.

\begin{table}[h]
\fontsize{8}{9}\selectfont 
\vspace{-8px}
\centering
\caption{Results on our dataset}
\label{tab:our_db}
\begin{tabular}{lcccc}
\textbf{Object} & \textbf{\cite{tejani2014latent}} & \textbf{\cite{brachmann2014learning}} & \multicolumn{1}{l}{\textbf{Our}} & \textbf{\begin{tabular}[c]{@{}c@{}}Our\\ joint optim.\end{tabular}} \\ \hline
\multicolumn{5}{c}{\rule{0pt}{3ex} \textbf{scenario 1 (supermarket objects)}}                                                                                            \\
amita     & 26.9             & 60.8             & \textbf{64.3}                             & 71.2                                                                   \\
colgate   & 22.8             & 11.1             & \textbf{26.1}                                & 28.6                                                                   \\
elite     & 10.1             & 71.9             & \textbf{74.9}                                & 77.6                                                                   \\
lipton    & 10.5             & 26.9             & \textbf{56.4}                                & 59.2                                                                   \\
oreo      & 26.9             & 44.4             & \textbf{58.5}                                & 59.3                                                                   \\
softkings & 26.3             & 26.9             & \textbf{75.5}                                & 75.9                                                                   \\
\multicolumn{5}{c}{\rule{0pt}{3ex} \textbf{scenario 2 (bin picking)}}                                                                                                    \\
coffeecup & 31.4                 & -               & 33.5                                  & 36.1                                                                     \\
juice     & 24.8                 & -               & 25.1                                  & 29                                                                   
\end{tabular}
\vspace{-6px}
\end{table}

\begin{figure}[t!]
\begin{center}
\includegraphics[width=0.9\linewidth]{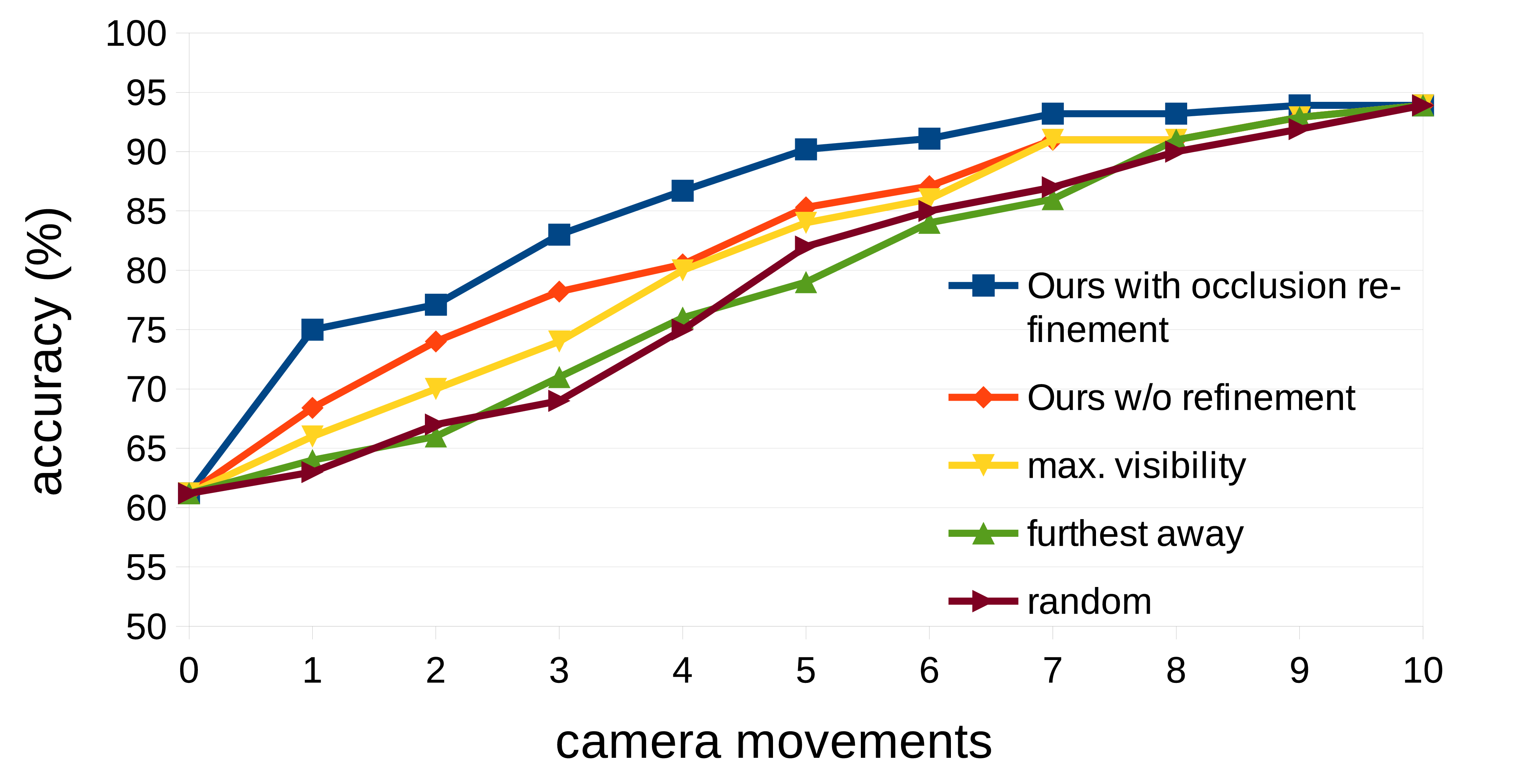}
\end{center}
\vspace{-10px}
   \caption{Results on active vision on our crowded dataset scenes}
\label{fig:active_res}
\vspace{-15px}
\end{figure}

\noindent \textbf{Active Vision Evaluation} We tested our active vision method on our dataset, using two different types of scenes. One is the crowded scenario used for single-shot evaluation, and the other depicts a special arrangement of objects, one behind the other in rows, that is commonly seen in a warehouse (Fig. \ref{fig:occ_ref}). All results takes into account all the object hypotheses during the next-best-view estimation. We compare our next-best-view prediction with and without occlusion refinement with three other baselines \cite{3dShap_Net}: a) maximum visibility (selecting a view that maximizes the visible area of the objects), b) furthest away (move the camera to the furthest point from all previous camera positions), c) move the camera randomly.

In the crowded scenario, we move the camera 10 times, measuring in each view the average pose estimation accuracy of the objects present in the scene (Fig. \ref{fig:active_res}). We see that our method without occlusion refinement slightly outperforms the maximum visibility baseline because usually the maximum reduction in entropy occurs when there is maximum visibility. Using occlusion refinement, however, we get a much better estimation of the entropy that is depicted in the performance.

\begin{figure}[t!]
\hspace{22px}
\includegraphics[width=0.8\linewidth]{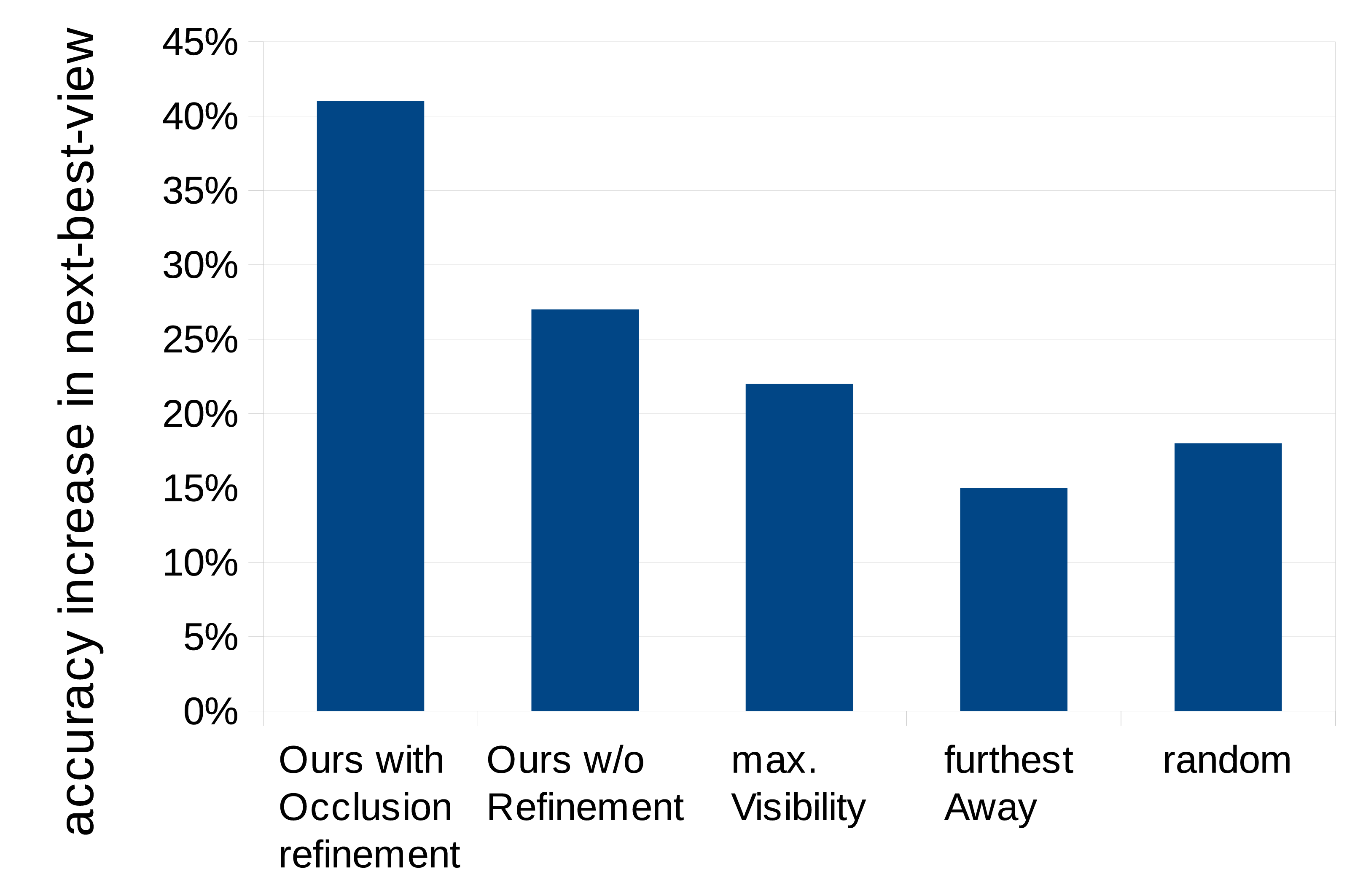}
   \caption{Results on active vision on scenes with objects arranged}
\label{fig:active_res2}
\vspace{-19px}
\end{figure}

When the objects are specially arranged, we were interested in measuring the increase in accuracy only in the single next-best-view, i.e. we allow the camera to move only once for speed reasons. This experiment (Fig. \ref{fig:active_res2}) makes very clear the importance of tackling occlusion when estimating the expected entropy. Our method with occlusion refinement was consistently finding the most appropriate view, whereas without this step, the next-best-view was usually the front view, with the objects behind being occluded.

Regarding the computational complexity of our single shot approach, training 3 layers of 800 features with $10^4$ patches for 100 epochs takes about 10mins on GPU. Our forest was trained with a larger set of $5 \cdot 10^6$ patches.
Thanks to our parallel implementation, we train a tree on an i7 CPU in 90 mins, while \cite{tejani2014latent} and \cite{brachmann2014learning} require about 3 and 1 days, respectively.
During testing, the main bottleneck is the Hough voting and mode extraction that takes about 4-7secs to execute, with an additional 2secs if joint optimization is used for 6 objects. Other methods need about 1sec.
\vspace{-4pt}

\vspace{-2pt}
\section{Conclusions}
\vspace{-4pt}
In this paper we proposed a complete framework for 6D object detection in crowded scenes, comprising of an unsupervised feature learning phase, 6 DoF object pose estimation using Hough Forests and a method for estimating the next-best-view using the trained forest. We conducted extensive evaluation on challenging public datasets, including a new one depicting realistic scenarios, using various state of the art methods. Our framework showed superior results, being able to generalize well to a variety of objects and scenes. As a future work, we want to investigate how different patch sizes can be combined, and explore how convolutional networks can help in this direction.

\noindent\textbf{Acknowledgment} The major part of the work was undertaken when A. Doumanoglou was at Imperial College London within the frames of his Ph.D. A. Doumanoglou was partially funded from the EU Horizon 2020 projects: RAMCIP (grant No 643433) and SARAFun (grant No 644938).

\bibliographystyle{ieee}
\bibliography{main_arxiv}

\end{document}